\documentclass[10pt,twocolumn,letterpaper]{article}

\usepackage[datasets]{wacv}      
\usepackage{graphicx}
\usepackage{subcaption}
\usepackage{float}
\usepackage{adjustbox}
\usepackage{graphicx}
\usepackage{tikz}
\usepackage{makecell}
\usepackage{array}
\usepackage{tikz}
\usepackage{bm}

\usetikzlibrary{
    decorations.pathreplacing,
    calc
}

\usepackage{booktabs}
\usepackage{multirow}

\usepackage{tabularx}
\usepackage{makecell}

\definecolor{wacvblue}{rgb}{0.21,0.49,0.74}
\usepackage[pagebackref,breaklinks,colorlinks,allcolors=wacvblue]{hyperref}

\def\wacvPaperID{2278} 
\def\confName{WACV}
\def\confYear{2027}

\title{Different Defects, Different Effects: A Controlled Audit of Face Recognition Training Data}

\author{
Ana Dias$^{1,2,3}$, João Ribeiro Pinto$^{4}$, Hugo Proença$^{1,2}$, João C. Neves$^{1,3}$\\[4pt]
\small $^{1}$University of Beira Interior, Portugal \quad $^{2}$IT: Instituto de Telecomunicações \quad $^{3}$NOVA LINCS \quad $^{4}$Amadeus, Portugal\\
\small ana.margarida.dias@ubi.pt\\
}

\begin{document}
\maketitle

\begin{abstract}

The development of face recognition has been strongly supported by the availability of large-scale training datasets. Among them, MS-Celeb-1M has played an important role by providing millions of facial images from nearly 100K identities. However, the size and heterogeneity of such datasets increase the likelihood of annotation errors, redundant samples, and other forms of data inconsistency. This motivated the introduction of refined versions, such as MS1MV2 and MS1MV3, aimed at improving the quality of the training data. Despite these cleaning efforts, we show that several types of training-data defects remain in MS1MV3. By proposing complementary data-auditing strategies, we identify five defect categories: exact and near-duplicate images, split identities, crop errors, multiple-face ambiguity, and invalid content. We then assess the effect of correcting these defects under a controlled face-recognition training protocol based on IResNet-50 and ArcFace. The results show that targeted interventions can preserve, and in some cases improve, recognition performance while requiring fewer optimization steps to reach the performance of the original training set. Duplicate removal provides the clearest overall benefit, while split-identity merging and invalid-content removal also improve training efficiency. Overall, the results indicate that the effect of dataset cleaning depends strongly on the type of defect being addressed and provide further evidence that training-data quality remains an important and comparatively underexplored factor in large-scale face recognition. The training manifests and code used in this work will be released upon acceptance.

\end{abstract}

\section{Introduction}

Modern face-recognition systems~\cite{arcface,adaface,magface} are commonly trained on large web-collected datasets~\cite{ms1m,vggface2} that provide the diversity needed for robust representation learning but make careful dataset-quality validation difficult. As a result, training data may contain defects that are often grouped as generic data-quality problems despite introducing different forms of error: duplicate images add repeated information, split identities create inconsistent supervision, and image-validity defects include crop errors, multiple-face ambiguity, and invalid content. Treating these distinct defects simply as ``noise'' can hide differences in how they affect recognition performance and training behavior.

\begin{figure}[t]
    \centering
    \includegraphics[
        width=\linewidth
    ]{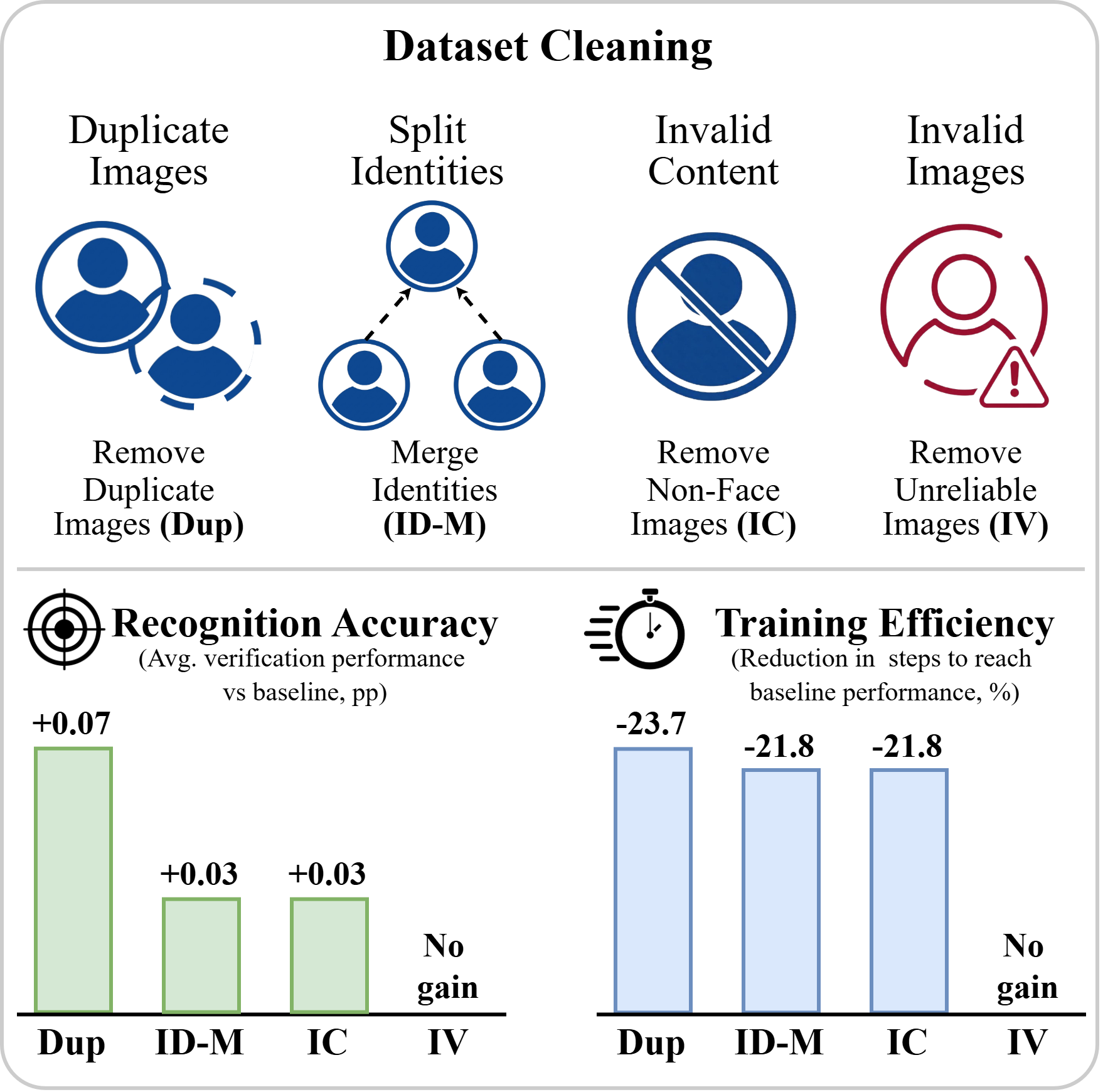}
    \vspace{-1.5em}
\caption{\textbf{Overview of the main training interventions and their effects.} Duplicate removal (Dup), split-identity merging (ID-M), invalid-content removal (IC), and image-validity filtering (IV) are compared with the baseline, showing that different cleaning interventions have distinct effects on recognition performance and training efficiency.}
    \label{fig:teaser}
    \vspace{-1.0em}
\end{figure}

Prior work has addressed training-data quality through data cleaning, noise-robust training, and redundancy reduction. Data-cleaning methods identify noisy samples and inconsistent labels~\cite{datacleaning,globallocal,adaptativenoise,webface260m}, noise-robust approaches reduce the influence of noisy supervision during training~\cite{subcenter_arcface}, and sample-selection and pruning strategies address training-data redundancy~\cite{facenms,diffprob}. However, these lines of work do not investigate which errors remain in curated datasets such as MS1MV3 or directly compare how correcting different dataset defects affects recognition performance and training behavior, leaving open whether different interventions provide similar benefits or whether their effects depend on the defect type.


In this work, we perform a controlled audit of MS1MV3~\cite{ms1mv3} and study five types of training-data defects: exact and near-duplicate images, split identities, crop errors, multiple-face ambiguity, and invalid content. Duplicate images are identified using perceptual hashes, while split identities are detected using identity-level face representations followed by image-level verification. Crop errors are detected when image padding overlaps the face or facial landmarks, multiple-face ambiguity is measured using the OFIQ~\cite{ofiq} SingleFacePresent criterion, and invalid content is identified with BLIP~\cite{BLIP} visual question answering.

The detected defects are translated into controlled training interventions. We evaluate duplicate removal at several thresholds, correct split identities by either merging their labels or removing the affected identity folders, and compare invalid-content removal with image-validity filtering. For the best-performing duplicate-removal setting, we additionally construct a random-removal control matched in total and per-identity image counts. We also evaluate selected combinations of the most promising individual interventions. This allows us to study the effects of different defect types separately, rather than producing a single cleaned version of the dataset. All variants are evaluated by training an IResNet-50 with ArcFace~\cite{arcface} under the same protocol. We consider both recognition performance and the number of optimization steps required to reach the best validation performance of the corresponding baseline run. Held-out verification is further evaluated on IJB-B and IJB-C~\cite{IJB}, comparing the validation-selected baseline with intervention models at earlier checkpoints.

Our results show that the effect of dataset cleaning depends strongly on the defect being corrected, as summarized in Figure~\ref{fig:teaser}. Duplicate removal provides the clearest overall benefit. The best-performing setting removes $15.44\%$ of the training images while increasing average validation accuracy from $97.20{\pm}0.02$ to $97.27{\pm}0.01$, and reaches the baseline performance with $23.7{\pm}5.0\%$ fewer optimization steps. A matched random-removal control with the same dataset size and per-identity image counts reaches only $97.18{\pm}0.01$, suggesting that the result is not explained by dataset reduction alone. Merging split identities largely preserves recognition accuracy while reaching the baseline performance with $21.8{\pm}5.0\%$ fewer steps and retaining all training images. Invalid-content removal shows a similar reduction in steps, while jointly filtering invalid content, crop errors, and multiple-face ambiguity does not show the same benefit. Combining identity merging with duplicate removal preserves validation performance and reaches the baseline performance with $24.3{\pm}2.9\%$ fewer steps, while extending the cleaning pipeline further does not provide an additional improvement. On the held-out IJB-B and IJB-C benchmarks, the strongest validation variants remain competitive with the baseline at earlier checkpoints, supporting the training-efficiency findings.

Our main contributions are:
\begin{itemize}
    \item We present a controlled audit of MS1MV3~\cite{ms1mv3} covering duplicate images, split identities, crop errors, multiple-face ambiguity, and invalid content, and design separate training interventions for the different defect types;
    \item We evaluate these interventions under a fixed face-recognition training protocol, considering recognition performance, the number of training steps required to reach the baseline performance, and held-out verification on IJB-B and IJB-C~\cite{IJB};
    \item We show that the effect of dataset cleaning is defect-dependent: duplicate removal provides the clearest overall benefit, while split-identity and image-validity interventions produce different effects on recognition performance and training efficiency.
\end{itemize}

\section{Related Work}

\subsection{Training-Data Quality in Face Recognition}

The performance of deep face recognition depends strongly on the scale and quality of the training data. Large-scale web-collected datasets such as MS-Celeb-1M~\cite{ms1m} and VGGFace2~\cite{vggface2} have become central to training modern face-recognition models, but their scale makes extensive manual verification impractical, motivating automatic and semi-automatic dataset cleaning~\cite{datacleaning}. 

A recurring challenge in these datasets is maintaining correct and consistent identity labels, since web-collected samples may contain both mislabeled images and multiple labels corresponding to the same person. One line of work addresses this problem by explicitly identifying and correcting errors in the training data. Wang et al.~\cite{devilnoise} studied the effect of label noise by distinguishing label flips from outliers. Global-Local GCN~\cite{globallocal} used cascaded graph convolutional networks to identify noisy samples and subsequently merge identity classes with similar feature centers, while Adaptive Label Noise Cleaning~\cite{adaptativenoise} used meta-supervision and class-adaptive thresholds to identify and remove noisy samples. WebFace260M~\cite{webface260m} introduced CAST, a self-training pipeline that iteratively used recognition models to clean WebFace260M and construct WebFace42M.

A separate line of work instead focuses on reducing the impact of noisy supervision during training. Hu et al.~\cite{noisetolerant} weighted training samples according to their estimated probability of being correctly labeled, Co-Mining~\cite{comining} used two peer networks to identify and exchange high-confidence clean samples during training, and Sub-center ArcFace~\cite{subcenter_arcface} represented each class with multiple sub-centers to improve robustness to label noise. Similarly, Ma et al.~\cite{rethinking} considered fine-grained intra- and inter-class conflicts and proposed evolving sub-centers to handle cases where the same identity appears under different class labels.

Beyond label consistency, prior work has also considered the suitability of individual face images for recognition. FaceQNet~\cite{faceqnet}, SER-FIQ~\cite{serfiq}, and CR-FIQA~\cite{crfiqa} explicitly estimate face-image quality, while MagFace~\cite{magface} learns face representations whose feature magnitude also provides a measure of sample quality.

In contrast, we study specific dataset defects at both the identity and image levels. We consider \emph{split identities}, where images of the same person are assigned to different training classes, and compare label merging with identity-folder removal. Among image-validity defects, we examine crop errors, multiple-face ambiguity, and invalid content while retaining challenging but valid variations such as pose, illumination, age, blur, and occlusion.

\subsection{Training-Data Redundancy in Face Recognition}

Large-scale face datasets may also contain substantial redundancy, particularly in the form of exact or near-duplicate images. Schlett et al.~\cite{doubletrouble} investigated how deduplication changes face-recognition and quality-assessment results produced by pretrained models. Our work instead investigates the effect of duplicate removal during model training, measuring both the resulting recognition performance and training efficiency.

Redundancy has also been addressed from the broader perspective of training-data reduction. Face-NMS~\cite{facenms} selected a sparse subset of face images in feature space by removing highly similar samples and retaining those with lower similarity to their class center, while DiffProb~\cite{diffprob} pruned samples with similar prediction probabilities within each identity and removed samples misclassified by a pretrained model.

Unlike these works, which address redundancy through model-dependent sample selection, we isolate image-level perceptual duplication as a specific training-data defect and retain one representative from each detected duplicate group. We further compare targeted duplicate removal with matched random removal, allowing us to distinguish the effect of eliminating redundancy from that of simply reducing the number of training images.



\newlength{\qualTopImg}
\setlength{\qualTopImg}{0.098\textwidth}

\newlength{\qualBottomImg}
\setlength{\qualBottomImg}{0.098\textwidth}

\newlength{\qualImgGap}
\setlength{\qualImgGap}{2pt}

\newlength{\qualPanelGap}
\setlength{\qualPanelGap}{0.02\textwidth}

\newlength{\qualRowGap}
\setlength{\qualRowGap}{10pt}

\newlength{\qualTitleGap}
\setlength{\qualTitleGap}{2pt}

\newlength{\qualGroupGap}
\setlength{\qualGroupGap}{1pt}

\newlength{\qualGroupSep}
\setlength{\qualGroupSep}{3pt}


\begin{figure*}[!h]
\centering


\makebox[\textwidth][c]{%

\begin{minipage}[t]{0.36\textwidth}
\centering

\textbf{Duplicate images}

\vspace{\qualTitleGap}

{\scriptsize Duplicate group 1}

\vspace{\qualGroupGap}

\begin{tabular}{
    @{}
    c
    @{\hspace{\qualImgGap}}
    c
    @{\hspace{\qualImgGap}}
    c
    @{}
}
\includegraphics[width=\qualTopImg]{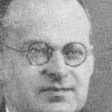} &
\includegraphics[width=\qualTopImg]{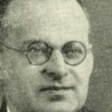} &
\includegraphics[width=\qualTopImg]{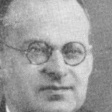}
\end{tabular}

\vspace{\qualGroupSep}

{\scriptsize Duplicate group 2}

\vspace{\qualGroupGap}

\begin{tabular}{
    @{}
    c
    @{\hspace{\qualImgGap}}
    c
    @{\hspace{\qualImgGap}}
    c
    @{}
}
\includegraphics[width=\qualTopImg]{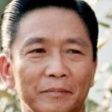} &
\includegraphics[width=\qualTopImg]{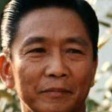} &
\includegraphics[width=\qualTopImg]{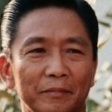}
\end{tabular}

\end{minipage}

\hspace{\qualPanelGap}

\begin{minipage}[t]{0.36\textwidth}
\centering

\textbf{Split identity}

\vspace{\qualTitleGap}

{\scriptsize Original identity A}

\vspace{\qualGroupGap}

\begin{tabular}{
    @{}
    c
    @{\hspace{\qualImgGap}}
    c
    @{\hspace{\qualImgGap}}
    c
    @{}
}
\includegraphics[width=\qualTopImg]{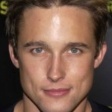} &
\includegraphics[width=\qualTopImg]{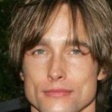} &
\includegraphics[width=\qualTopImg]{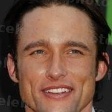}
\end{tabular}

\vspace{\qualGroupSep}

{\scriptsize Original identity B}

\vspace{\qualGroupGap}

\begin{tabular}{
    @{}
    c
    @{\hspace{\qualImgGap}}
    c
    @{\hspace{\qualImgGap}}
    c
    @{}
}
\includegraphics[width=\qualTopImg]{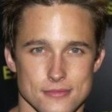} &
\includegraphics[width=\qualTopImg]{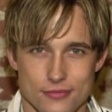} &
\includegraphics[width=\qualTopImg]{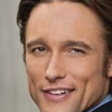}
\end{tabular}

\end{minipage}
}

\vspace{\qualRowGap}


\makebox[\textwidth][c]{%

\begin{minipage}[t]{0.29\textwidth}
\centering

\textbf{Crop error}

\vspace{\qualTitleGap}

\begin{tabular}{
    @{}
    c
    @{\hspace{\qualImgGap}}
    c
    @{\hspace{\qualImgGap}}
    c
    @{}
}
\includegraphics[width=\qualBottomImg]{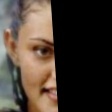} &
\includegraphics[width=\qualBottomImg]{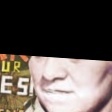} &
\includegraphics[width=\qualBottomImg]{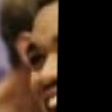}
\\[\qualImgGap]
\includegraphics[width=\qualBottomImg]{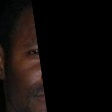} &
\includegraphics[width=\qualBottomImg]{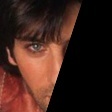} &
\includegraphics[width=\qualBottomImg]{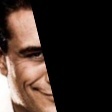}
\end{tabular}

\end{minipage}

\hspace{\qualPanelGap}

\begin{minipage}[t]{0.29\textwidth}
\centering

\textbf{Multiple-face ambiguity}

\vspace{\qualTitleGap}

\begin{tabular}{
    @{}
    c
    @{\hspace{\qualImgGap}}
    c
    @{\hspace{\qualImgGap}}
    c
    @{}
}
\includegraphics[width=\qualBottomImg]{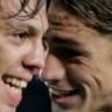} &
\includegraphics[width=\qualBottomImg]{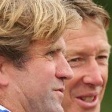} &
\includegraphics[width=\qualBottomImg]{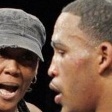}
\\[\qualImgGap]
\includegraphics[width=\qualBottomImg]{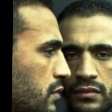} &
\includegraphics[width=\qualBottomImg]{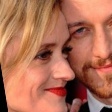} &
\includegraphics[width=\qualBottomImg]{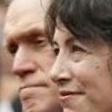}
\end{tabular}

\end{minipage}

\hspace{\qualPanelGap}

\begin{minipage}[t]{0.29\textwidth}
\centering

\textbf{Invalid content}

\vspace{\qualTitleGap}

\begin{tabular}{
    @{}
    c
    @{\hspace{\qualImgGap}}
    c
    @{\hspace{\qualImgGap}}
    c
    @{}
}
\includegraphics[width=\qualBottomImg]{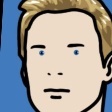} &
\includegraphics[width=\qualBottomImg]{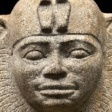} &
\includegraphics[width=\qualBottomImg]{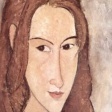}
\\[\qualImgGap]
\includegraphics[width=\qualBottomImg]{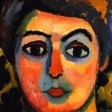} &
\includegraphics[width=\qualBottomImg]{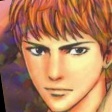} &
\includegraphics[width=\qualBottomImg]{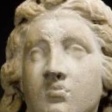}
\end{tabular}

\end{minipage}
}


\vspace{-0.5em}

\caption{\textbf{Representative examples of the dataset defects considered in this work.} Duplicate images are shown in two groups, while the split-identity examples show images originally assigned to different identity labels but identified as belonging to the same person. The remaining panels show examples of crop errors, multiple-face ambiguity, and invalid content.}
\label{fig:defect_examples}

\vspace{-1em}

\end{figure*}

\section{Dataset Defect Detection}

We consider five types of face-dataset defects: split identities, exact and near-duplicate images, crop errors, multiple-face ambiguity, and invalid content. These include both image-level defects, affecting the validity or redundancy of individual samples, and identity-level defects, affecting identity-label consistency. Valid but challenging facial variations, including pose, severe blur, and occlusion, are retained. Figure~\ref{fig:defect_examples} provides examples of the considered cases.

\subsection{Split-Identity Detection}
\label{sec:identity_cleaning}

A split identity occurs when images of the same person are assigned to different identity labels. We identify candidate identity pairs using face embeddings and verify them through cross-identity image comparisons.

\vspace{-0.8em}
\paragraph{Identity representations.}
For each identity \(i\), we construct a representative centroid from the face embeddings of its images. Let \(\mathbf{e}_{i,k}\in\mathbb{R}^{d}\) denote the \(\ell_2\)-normalized embedding of image \(k\) from identity \(i\). An initial centroid is obtained by averaging all embeddings of that identity. The images are then ranked according to the cosine similarity of their embeddings to this centroid. To reduce the influence of less representative samples, only the top \(80\%\) most centroid-consistent embeddings are retained in the set \(\mathcal{T}_i\). The final identity representation is obtained by averaging these embeddings,

\begin{equation}
    \bar{\mathbf{e}}_i =
    \frac{1}{|\mathcal{T}_i|}
    \sum_{k\in\mathcal{T}_i}\mathbf{e}_{i,k},
\end{equation}

and $\ell_2$-normalizing the resulting vector to obtain the final centroid $\mathbf{c}_i$.

\vspace{-0.8em}
\paragraph{Candidate split-identity search.}
For each identity \(i\), we retrieve the \(N\) other identities with the most similar centroids using HNSW~\cite{HNSW}. Each retrieved identity \(j\) forms a candidate pair \((i,j)\). The cosine similarity between their centroids is then measured, and the pair is retained only when

\begin{equation}
    \operatorname{cos}\!\left(\mathbf{c}_i,\mathbf{c}_j\right) \geq \tau_c,
\end{equation}

where \(\tau_c\) denotes the centroid-similarity threshold. The retained pairs correspond to identity folders that are sufficiently similar to be considered possible splits of the same person.

\vspace{-0.8em}
\paragraph{Cross-folder verification.}
The centroid-based search identifies identity pairs that are similar at the folder level, but this alone is not sufficient to conclude that they contain the same person. We therefore require supporting matches from multiple nonredundant images in both folders.

For each retained pair \((i,j)\), the images in each folder are ranked by their similarity to the corresponding identity centroid,

\begin{equation}
    s_{i,k} = \operatorname{cos}\!\left(\mathbf{e}_{i,k},\mathbf{c}_i\right).
\end{equation}

From the top \(80\%\) most centroid-consistent images, we select up to 20 representatives per folder. An image is skipped if its cosine similarity to any previously selected representative from the same folder is at least \(0.90\). This avoids treating repeated or highly similar images as independent evidence. The resulting representative sets \(\mathcal{R}_i\) and \(\mathcal{R}_j\) may therefore contain fewer than 20 images.

The two representative sets are then compared. Each image in \(\mathcal{R}_i\) is matched to its most similar image in \(\mathcal{R}_j\), and it provides support when this similarity is at least \(0.60\). The same procedure is applied in the opposite direction. We require multiple supporting images from each folder, and the criterion must be satisfied independently in both directions. In addition, the supporting images from each folder must correspond to at least two distinct images in the opposite folder, preventing several images from confirming the pair through a single counterpart.


\paragraph{Identity grouping.}
Verified identities are grouped conservatively using a complete-link requirement: every pair of identities within a group must pass cross-folder verification. Overlapping candidate groups that do not satisfy this requirement are resolved by prioritizing pairs with stronger similarity evidence. A random sample of detected split-identity pairs was manually inspected to assess detection precision. The validation results are reported in the supplementary material.

\subsection{Duplicate Image Detection}
\label{sec:duplicate_detection}

Duplicate detection is performed independently within each identity, so images assigned to different identities are not compared at this stage. For each image \(x_i\), a perceptual hash \(h_p(x_i)\) is obtained. Two images are considered duplicate candidates when the Hamming distance between their perceptual hashes does not exceed a threshold \(\tau_h\),

\begin{equation}
    d_H\!\left(h_p(x_i),h_p(x_j)\right) \leq \tau_h.
\end{equation}

This definition includes identical perceptual hashes at distance zero, allowing exact and near-duplicate images to be handled within the same procedure.

We group duplicate candidates transitively. If one image matches a second image and the second matches a third, all three are assigned to the same duplicate group, even if the first and third images do not directly match. This allows all images connected through duplicate relations to be treated as a single group.

\subsection{Image-Validity Defect Detection}

The remaining defects concern image validity and include crop errors, multiple-face ambiguity, and invalid content. Each category is identified using a dedicated detection criterion.

\vspace{-1em}
\paragraph{Crop-error detection.}
Some aligned face crops contain synthetic black padding introduced when the original face region extends beyond the available image area. Padding alone is not considered an error, since it may occur without affecting the visible face. We therefore identify crop errors based on how much detected padding overlaps the face.

Black regions connected to the image border are first identified as candidate padding.
The largest detected face and its facial landmarks are then used for the crop-error assessment. Padding is measured within the face bounding box and within a central facial region defined by the eye, nose, and mouth landmarks.
An image is marked as a crop error when the padding covers at least \(10\%\) of the central facial region, at least \(20\%\) of the face bounding box, or at least \(10\%\) of the face bounding box while also covering two or more facial landmarks.

\vspace{-1em}
\paragraph{Multiple-face ambiguity detection.}
Some images contain more than one visible face, but small background faces do not necessarily make the intended identity ambiguous. We therefore use the OFIQ~\cite{ofiq} SingleFacePresent measure to quantify the relative size of the second-largest detected face,

\begin{equation}
    r_{\mathrm{face}} = \frac{A_{\mathrm{second}}}{A_{\mathrm{largest}}},
\end{equation}

where \(A_{\mathrm{second}}\) and \(A_{\mathrm{largest}}\) denote the bounding-box areas of the second-largest and largest detected faces, respectively. An image is marked as ambiguous when \(r_{\mathrm{face}} > 0.15\), indicating that the secondary face is sufficiently prominent to introduce ambiguity about the intended identity.

\paragraph{Invalid-content detection.}
Some samples do not correspond to photographs of real people and instead contain content such as cartoons, drawings, paintings, or statues. We identify these cases using BLIP~\cite{BLIP} visual question answering. For each image, BLIP is queried with the questions ``Is this a photograph of a real person?'', ``Is this a cartoon?'', ``Is this a drawing?'', ``Is this a painting?'', and ``Is this a statue?''

An image is marked as invalid content only when BLIP gives a negative response to the real-person photograph question and a positive response for at least one of the four non-real categories. A random sample of invalid-content detections was manually inspected to assess detection precision. The validation results are reported in the supplementary material.

\section{Experimental Setup}

\subsection{Dataset and Controlled Interventions}
\label{sec:dataset_interventions}

\begin{table*}[t]
\centering
\caption{\textbf{Dataset interventions, resulting training-set sizes, and distribution of removals across identities.} All reductions are measured relative to Orig. \emph{IDs Retained} denotes the number of identity labels remaining after each intervention, while \emph{IDs with Removals (\%)} is the percentage of original identities from which at least one image is removed. Considering only the affected identities, \emph{Removed Images} reports the mean / median number of removed images per identity, while \emph{Removal Rate (\%)} reports the mean / median percentage of images removed per affected identity. Crop and Multi-Face are included for completeness but are not evaluated as standalone training conditions.}
\label{tab:interventions}

\footnotesize
\renewcommand{\arraystretch}{1.15}
\setlength{\tabcolsep}{3pt}

\begin{adjustbox}{max width=\textwidth}
\begin{tabular}{
    @{}
    >{\raggedright\arraybackslash}m{0.095\textwidth}
    >{\raggedright\arraybackslash}m{0.265\textwidth}
    >{\centering\arraybackslash}m{0.095\textwidth}
    >{\centering\arraybackslash}m{0.090\textwidth}
    >{\centering\arraybackslash}m{0.075\textwidth}
    >{\centering\arraybackslash}m{0.100\textwidth}
    >{\centering\arraybackslash}m{0.110\textwidth}
    >{\centering\arraybackslash}m{0.115\textwidth}
    @{}
}

\toprule

\makebox[\linewidth][c]{Condition}
&
\makebox[\linewidth][c]{%
    \shortstack[c]{Dataset\\modification}%
}
&
\shortstack[c]{Images\\Retained}
&
\shortstack[c]{Image Reduction\\(\%)}
&
\shortstack[c]{IDs\\Retained}
&
\shortstack[c]{IDs with\\Removals (\%)}
&
\multicolumn{2}{c}{%
    \begin{tabular}{@{}cc@{}}
        \multicolumn{2}{c}{Per affected identity} \\[-0.3ex]
        \cmidrule(lr){1-2}
        \shortstack[c]{Removed images\\(mean / median)}
        &
        \shortstack[c]{Removal rate (\%)\\(mean / median)}
    \end{tabular}%
}
\\

\midrule

Orig.
& Original dataset
& $5{,}179{,}510$
& $0.00$
& $93{,}431$
& --
& --
& -- \\

\midrule

D5
& Deduplication, $\tau_h=5$
& $4{,}575{,}329$
& $11.66$
& $93{,}431$
& $83.23$
& $7.77 / 4$
& $12.89 / 8.33$ \\

D7
& Deduplication, $\tau_h=7$
& $4{,}379{,}554$
& $15.44$
& $93{,}431$
& $88.70$
& $9.65 / 6$
& $16.27 / 11.54$ \\

R-D7
& Random removal matched to D7
& $4{,}379{,}554$
& $15.44$
& $93{,}431$
& $88.70$
& $9.65 / 6$
& $16.27 / 11.54$ \\

D9
& Deduplication, $\tau_h=9$
& $4{,}155{,}394$
& $19.77$
& $93{,}431$
& $91.67$
& $11.96 / 8$
& $19.91 / 15.25$ \\

\midrule

ID-M
& Split-identity merging
& $5{,}179{,}510$
& $0.00$
& $91{,}468$
& --
& --
& -- \\

ID-D
& Split-identity folder removal
& $5{,}126{,}117$
& $1.03$
& $91{,}468$
& $2.10$
& $27.20 / 21$
& $100.00 / 100.00$ \\

\midrule

Crop
& Crop-error subset removed
& $5{,}151{,}099$
& $0.55$
& $93{,}430$
& $20.98$
& $1.45 / 1$
& $3.33 / 2.20$ \\

Multi-Face
& Multiple-face subset removed
& $5{,}167{,}184$
& $0.24$
& $93{,}428$
& $8.30$
& $1.59 / 1$
& $3.13 / 2.17$ \\

IC
& Invalid-content subset removed
& $5{,}128{,}572$
& $0.98$
& $93{,}153$
& $22.95$
& $2.38 / 1$
& $7.33 / 2.78$ \\

IV
& All three image-validity subsets removed
& $5{,}088{,}730$
& $1.75$
& $93{,}147$
& $43.26$
& $2.25 / 1$
& $6.04 / 3.03$ \\

\midrule

ID-M$\rightarrow$D7
& Identity merge $\rightarrow$ deduplication at $\tau_h=7$
& $4{,}357{,}935$
& $15.86$
& $91{,}468$
& $89.38$
& $9.84 / 6$
& $16.92 / 11.76$ \\

IC$\rightarrow$ID-M$\rightarrow$D7
& Invalid content $\rightarrow$ identity merge $\rightarrow$ deduplication at $\tau_h=7$
& $4{,}318{,}991$
& $16.61$
& $91{,}203$
& $90.77$
& $10.15 / 6$
& $18.00 / 12.50$ \\

\bottomrule
\end{tabular}
\end{adjustbox}
\end{table*}

\newcommand{\hdr}[1]{%
    \begin{tabular}[c]{@{}c@{}}#1\end{tabular}%
}

\begin{table}[t]
\centering
\caption{\textbf{Structure of the detected duplicate and split-identity groups.} Group members correspond to images for duplicate groups, and to original identity folders for split-identity groups.}
\label{tab:group_structure}

\small
\setlength{\tabcolsep}{4pt}
\renewcommand{\arraystretch}{1.05}

\resizebox{\columnwidth}{!}{%
\begin{tabular}{@{}lcccc@{}}
\toprule
\hdr{Condition}
& \hdr{Groups}
& \hdr{Group\\members}
& \hdr{Group size\\(mean / median)}
& \hdr{Max.\\group size} \\
\midrule

D5
& $374{,}485$
& $978{,}666$
& $2.61 / 2$
& $105$ \\

D7
& $473{,}525$
& $1{,}273{,}481$
& $2.69 / 2$
& $108$ \\

D9
& $554{,}419$
& $1{,}578{,}535$
& $2.85 / 2$
& $108$ \\

Split identities
& $1{,}887$
& $3{,}850$
& $2.04 / 2$
& $4$ \\

\bottomrule
\end{tabular}%
}
\end{table}

We use MS1MV3~\cite{ms1mv3} as the source training dataset, comprising 5,179,510 face images from 93,431 identities. Each detected defect is translated into a controlled training intervention, implemented through a separate manifest that specifies the training images and identity labels without modifying the source files. Table~\ref{tab:interventions} summarizes the interventions, resulting training-set sizes, and distribution of image removals across identities. Table~\ref{tab:group_structure} summarizes the structure of the detected duplicate and split-identity groups.

\vspace{-0.5em}

\paragraph{Individual interventions.}
\textbf{Original (Orig.)} corresponds to the unmodified MS1MV3~\cite{ms1mv3} training set. For duplicate removal, \textbf{D5}, \textbf{D7}, and \textbf{D9} retain one representative from each duplicate group detected with $\tau_h=5$, $7$, and $9$, respectively. The representative is selected as the image with the largest file size, with ties resolved by filename. For \textbf{D7}, we additionally construct a matched random-removal control (\textbf{R-D7}) that removes the same number of randomly selected images from each identity, preserving the resulting dataset size and per-identity image counts. Additional ablations comparing transitive and non-transitive grouping of duplicate images, as well as different representative-image selection strategies, are reported in the supplementary material.

For each verified split-identity group, the largest folder is selected as the canonical identity, with ties resolved by identity name. \textbf{ID-M} assigns all folders in the group a common training label while retaining all images, while \textbf{ID-D} retains only the canonical folder.

Crop errors, multiple-face ambiguity, and invalid content are quantified separately during the dataset audit. We use \textbf{invalid content} specifically for samples that do not correspond to photographs of real people. \textbf{IV} filters all three image-validity subsets, while \textbf{IC} filters invalid content alone. Crop and Multi-Face are not evaluated separately as training conditions, as they affect only a small fraction of the dataset.

\vspace{-0.5em}

\paragraph{Combined interventions.}
We additionally evaluate combinations of the strongest individual interventions. When multiple operations are combined, invalid-content removal is applied first, followed by identity merging and duplicate removal. Identity merging precedes duplicate removal so that duplicates across previously separate labels can also be detected.

\subsection{Implementation Details}

Duplicate detection uses the PHash implementation from \texttt{imagededup}~\cite{imagededup}. For split-identity detection, face embeddings are extracted using the InsightFace~\cite{insightface} \texttt{buffalo\_l} recognition model. Since MS1MV3~\cite{ms1mv3} already contains aligned face crops, no additional face detection or alignment is performed during embedding extraction. The resulting embeddings are \(\ell_2\)-normalized before obtaining identity representations. Candidate split identities are retrieved using HNSW~\cite{HNSW} with \(N=5\), followed by a centroid-similarity threshold of \(\tau_c=0.70\). For cross-folder verification, the required support in each direction is \(\max(2,\lceil 0.15m_i\rceil)\), where \(m_i\) is the number of selected representatives, with at least two distinct matched counterparts required. The split-identity parameters were fixed before the training experiments following qualitative inspection of candidate matches. 

Crop-error detection uses InsightFace~\cite{insightface} face detection and 106-point facial landmarks. Multiple-face ambiguity is assessed using the OFIQ~\cite{ofiq} SingleFacePresent measure, and invalid-content detection uses the BLIP-VQA base model~\cite{BLIP}.

\begin{table*}[t]
\centering
\caption{\textbf{Validation results averaged over two training seeds (mean $\pm$ std).} Best Step denotes the training step at which the best validation performance is achieved. Orig.-Match Step denotes the first training step at which a variant reaches the best validation performance of the original baseline, and Step reduction reports the corresponding relative reduction in training steps. Step-based statistics are reported only when the baseline is reached in both seeds. Bold and underlined values denote the best and second-best results, respectively.}
\label{tab:validation_results}
\footnotesize
\setlength{\tabcolsep}{3pt}
\renewcommand{\arraystretch}{1.15}

\begin{adjustbox}{max width=\textwidth}
\begin{tabular}{@{}lccccccccc@{}}
\toprule
Variant
& LFW
& AgeDB
& CALFW
& CPLFW
& CFP-FP
& Avg.
& Best Step
& Orig.-Match Step
& Step reduc. (\%) \\
\midrule

\textbf{Orig.}
& $99.82{\pm}0.02$
& $98.21{\pm}0.16$
& $96.10{\pm}0.03$
& $92.91{\pm}0.01$
& $98.92{\pm}0.11$
& $97.20{\pm}0.02$
& $208.0{\pm}7.1$k
& $208.0{\pm}7.1$k
& -- \\

\midrule
\multicolumn{10}{@{}c}{\textit{Duplicate removal}} \\[2pt]
\midrule

D5
& $99.85{\pm}0.03$
& $98.24{\pm}0.02$
& $96.10{\pm}0.03$
& $93.03{\pm}0.21$
& $98.86{\pm}0.10$
& $97.22{\pm}0.02$
& $182.5{\pm}17.7$k
& --
& -- \\

D7
& $99.83{\pm}0.00$
& $98.28{\pm}0.14$
& $96.13{\pm}0.04$
& $93.20{\pm}0.11$
& $98.90{\pm}0.05$
& $\mathbf{97.27{\pm}0.01}$
& $195.0{\pm}4.2$k
& \underline{$158.5{\pm}5.0$k}
& \underline{$23.7{\pm}5.0$} \\

R-D7
& $99.81{\pm}0.04$
& $98.20{\pm}0.10$
& $96.16{\pm}0.04$
& $92.88{\pm}0.08$
& $98.86{\pm}0.00$
& $97.18{\pm}0.01$
& $230.0{\pm}14.1$k
& --
& -- \\

D9
& $99.83{\pm}0.01$
& $98.15{\pm}0.11$
& $96.15{\pm}0.10$
& $93.12{\pm}0.02$
& $98.88{\pm}0.01$
& $97.22{\pm}0.00$
& $211.0{\pm}29.7$k
& $211.0{\pm}29.7$k
& $-1.3{\pm}10.8$ \\

\midrule
\multicolumn{10}{@{}c}{\textit{Split-identity interventions}} \\[2pt]
\midrule
ID-M
& $99.83{\pm}0.00$
& $98.24{\pm}0.06$
& $96.14{\pm}0.02$
& $93.05{\pm}0.11$
& $98.86{\pm}0.02$
& $97.23{\pm}0.04$
& $174.5{\pm}21.9$k
& $162.5{\pm}5.0$k
& $21.8{\pm}5.0$ \\

ID-D
& $99.84{\pm}0.01$
& $98.26{\pm}0.06$
& $96.14{\pm}0.05$
& $93.02{\pm}0.12$
& $98.85{\pm}0.06$
& $97.23{\pm}0.01$
& $184.0{\pm}24.0$k
& $184.0{\pm}24.0$k
& $11.3{\pm}14.6$ \\

\midrule
\multicolumn{10}{@{}c}{\textit{Image-validity interventions}} \\[2pt]
\midrule

IV
& $99.83{\pm}0.04$
& $98.23{\pm}0.06$
& $96.08{\pm}0.00$
& $92.96{\pm}0.01$
& $98.84{\pm}0.00$
& $97.19{\pm}0.01$
& $192.0{\pm}26.9$k
& --
& -- \\

IC
& $99.82{\pm}0.02$
& $98.30{\pm}0.10$
& $96.07{\pm}0.07$
& $93.08{\pm}0.04$
& $98.90{\pm}0.14$
& $97.23{\pm}0.04$
& $171.0{\pm}14.1$k
& $162.5{\pm}2.1$k
& $21.8{\pm}3.7$ \\

\midrule
\multicolumn{10}{@{}c}{\textit{Combined interventions}} \\[2pt]
\midrule

ID-M$\rightarrow$D7
& $99.79{\pm}0.06$
& $98.30{\pm}0.07$
& $96.16{\pm}0.01$
& $93.04{\pm}0.09$
& $98.96{\pm}0.02$
& \underline{$97.26{\pm}0.01$}
& $203.5{\pm}47.4$k
& $\mathbf{157.5{\pm}0.7}$k
& $\mathbf{24.3{\pm}2.9}$ \\


IC$\rightarrow$ID-M$\rightarrow$D7
& $99.84{\pm}0.01$
& $98.29{\pm}0.05$
& $96.10{\pm}0.03$
& $92.98{\pm}0.08$
& $99.03{\pm}0.00$
& $97.25{\pm}0.02$
& $229.0{\pm}21.2$k
& $168.0{\pm}22.6$k
& $19.0{\pm}13.6$ \\

\bottomrule
\end{tabular}
\end{adjustbox}
\vspace{-1.5em}
\end{table*}

\subsection{Training and Evaluation Protocol}

All variants use the same training protocol and differ only in the training manifest and, for identity merging, the class labels. As our face-recognition model, we train an IResNet-50 with 512-D embeddings using ArcFace~\cite{arcface} with margin \(m=0.5\) and scale \(s=64\). Images are resized to \(112\times112\) and randomly horizontally flipped with probability \(0.5\). Training uses SGD with batch size 512, learning rate \(0.1\), momentum \(0.9\), and weight decay \(5\times10^{-4}\). The learning rate is reduced by a factor of \(0.1\) at 80k and 140k steps, and training continues for 250k steps. Each variant is trained with two seeds.

Checkpoints are evaluated every 1,000 steps on LFW~\cite{lfw}, CALFW~\cite{calfw}, CPLFW~\cite{cplfw}, AgeDB~\cite{agedb}, and CFP-FP~\cite{cfp}. For each run, we select the checkpoint with the highest mean accuracy across the five benchmarks. Training efficiency is measured by the first step at which a variant reaches the best validation performance of the corresponding original run. Variants that do not reach this target within 250k steps are reported as not reaching it.

Held-out verification is additionally evaluated on IJB-B and IJB-C~\cite{IJB}, which are not used for checkpoint or intervention selection. The original baseline is evaluated at the step with its best validation performance, while the intervention variants are evaluated at earlier steps selected around baseline-matching performance. We report TAR at FARs of \(10^{-6}\), \(10^{-5}\), \(10^{-4}\), and \(10^{-3}\).

\section{Results}

\begin{table*}[t]
\centering
\caption{\textbf{Held-out verification performance on IJB-B and IJB-C, averaged over two training seeds (mean $\pm$ std).} The original baseline is evaluated at the step with its best validation performance, while the intervention variants are evaluated at earlier steps selected around baseline-matching performance. TAR is reported at FARs of $10^{-6}$, $10^{-5}$, $10^{-4}$, and $10^{-3}$. Bold values are equal to or above the original baseline, while underlined values are below the baseline by at most 0.05.}
\label{tab:ijb_results_avg_closest}
\vspace{-0.5em}

\footnotesize
\setlength{\tabcolsep}{2.1pt}
\renewcommand{\arraystretch}{1.15}

\begin{adjustbox}{max width=\textwidth}
\begin{tabular}{@{}lcccccccc@{}}
\toprule
& \multicolumn{4}{c}{IJB-B} & \multicolumn{4}{c}{IJB-C} \\
\cmidrule(lr){2-5}
\cmidrule(lr){6-9}
Variant
& $10^{-6}$
& $10^{-5}$
& $10^{-4}$
& $10^{-3}$
& $10^{-6}$
& $10^{-5}$
& $10^{-4}$
& $10^{-3}$ \\
\midrule

\textbf{Orig.}
& $38.27 \pm 0.72$
& $91.07 \pm 0.21$
& $95.06 \pm 0.02$
& $96.56 \pm 0.02$
& $90.39 \pm 0.09$
& $94.68 \pm 0.06$
& $96.39 \pm 0.01$
& $97.53 \pm 0.02$ \\

D7
& $\bm{39.48 \pm 0.41}$
& $90.34 \pm 0.23$
& $94.90 \pm 0.05$
& $96.50 \pm 0.03$
& $90.32 \pm 0.26$
& $94.42 \pm 0.00$
& $\underline{96.35 \pm 0.03}$
& $\bm{97.57 \pm 0.01}$ \\

ID-M
& $37.65 \pm 0.53$
& $\bm{91.14 \pm 0.45}$
& $94.90 \pm 0.02$
& $\bm{96.61 \pm 0.02}$
& $90.21 \pm 0.30$
& $94.57 \pm 0.08$
& $\underline{96.34 \pm 0.04}$
& $\bm{97.63 \pm 0.05}$ \\

IC
& $37.73 \pm 0.60$
& $\bm{91.50 \pm 0.34}$
& $94.84 \pm 0.03$
& $96.46 \pm 0.04$
& $90.21 \pm 0.26$
& $\underline{94.65 \pm 0.10}$
& $\underline{96.35 \pm 0.03}$
& $\underline{97.49 \pm 0.03}$ \\

ID-M$\rightarrow$D7
& $\bm{40.11 \pm 0.12}$
& $\bm{91.39 \pm 0.04}$
& $94.94 \pm 0.04$
& $\bm{96.57 \pm 0.03}$
& $\bm{90.62 \pm 0.26}$
& $\underline{94.63 \pm 0.07}$
& $\underline{96.36 \pm 0.01}$
& $\bm{97.53 \pm 0.00}$ \\

IC$\rightarrow$ID-M$\rightarrow$D7
& $37.61 \pm 0.80$
& $90.76 \pm 0.88$
& $\bm{95.10 \pm 0.06}$
& $\bm{96.56 \pm 0.04}$
& $90.16 \pm 0.54$
& $\bm{94.72 \pm 0.12}$
& $\bm{96.40 \pm 0.04}$
& $\bm{97.58 \pm 0.03}$ \\

\bottomrule
\end{tabular}
\end{adjustbox}
\vspace{-2.0em}
\end{table*}


\subsection{Recognition performance}
Table~\ref{tab:validation_results} summarizes recognition performance on the five validation benchmarks. Among the duplicate-removal variants, the $\tau_h=7$ setting (D7) achieves the highest average validation accuracy, $97.27$, while the matched random-removal control (R-D7) reaches $97.18$. Since R-D7 matches D7 in dataset size and per-identity image counts, this comparison suggests that the improvement is associated with targeted duplicate removal rather than image removal alone. The neighboring duplicate thresholds also do not show the same improvement, with both D5 ($\tau_h=5$) and D9 ($\tau_h=9$) reaching an average accuracy of $97.22$.

For split identities, split-identity merging (ID-M) and folder removal (ID-D) both reach an average accuracy of $97.23$. Because ID-M retains all images while matching the performance of ID-D, the result suggests that the label inconsistency can be addressed without discarding the affected images.

For image-validity defects, invalid-content removal (IC) reaches $97.23$, while image-validity filtering (IV), which additionally removes crop errors and multiple-face ambiguity, reaches $97.19$. This difference may reflect the nature of the defects: invalid content introduces samples that can interfere with identity learning, while crop errors and multiple-face ambiguity mainly introduce additional visual variability.

When the interventions are combined, their effects do not accumulate. Split-identity merging followed by D7 duplicate removal (ID-M$\rightarrow$D7) reaches $97.26$, closely matching D7 while addressing both label inconsistencies and duplicate images. Extending this pipeline with invalid-content removal (IC$\rightarrow$ID-M$\rightarrow$D7) reaches $97.25$, with no further increase in average validation accuracy. Across the interventions, targeted duplicate removal achieves the strongest result, while correcting additional defect types mainly preserves rather than further improves validation performance.

\subsection{Training Efficiency}

Table~\ref{tab:validation_results} also summarizes the number of training steps required to reach the best validation performance of the original run. Among the duplicate-removal variants, D7 shows a $23.7{\pm}5.0\%$ reduction in steps, while D5 reaches the target in only one seed and D9 shows no consistent reduction. The matched random-removal control (R-D7) also fails to reach the target in both seeds, reinforcing that the efficiency gain observed for D7 is not reproduced by dataset reduction alone.

For split identities, split-identity merging (ID-M) reduces the required steps by $21.8$, compared with $11.3$ for folder removal (ID-D). Together with their matched validation accuracy, these results support correcting the label inconsistency through merging rather than discarding the affected folders.

For image-validity defects, invalid-content removal (IC) reduces the steps by $21.8$, while image-validity filtering (IV) reaches the target in only one seed. This follows the pattern observed in validation accuracy, with invalid-content removal showing a clearer benefit than additionally filtering crop errors and multiple-face ambiguity.

Among the combined interventions, ID-M$\rightarrow$D7 provides the largest reduction, $24.3$. Extending the pipeline with invalid-content removal results in a smaller and more variable reduction of $19.0$, indicating that the efficiency gains of the individual interventions do not simply accumulate.

Across the experiments, the clearest efficiency gains are obtained with D7 duplicate removal, split-identity merging, invalid-content removal, and their ID-M$\rightarrow$D7 combination. As with recognition performance, these gains depend on the intervention being applied rather than on the amount of data removed or the number of defect types addressed.





\subsection{Held-Out IJB Evaluation}




Table~\ref{tab:ijb_results_avg_closest} evaluates intervention models at earlier checkpoints selected around baseline-matching performance on IJB-B and IJB-C. Despite using fewer training steps than the validation-selected original models, the interventions remain close to or outperform the baseline across many of the evaluated operating points. This is particularly clear on IJB-C at FARs of $10^{-4}$ and $10^{-3}$, where all evaluated interventions are either equal to, above, or within 0.05 of the original baseline.

The variants with the strongest validation results also perform well on IJB-B and IJB-C. D7 stays close to the original baseline overall, with gains at some operating points. ID-M shows stronger overall results on the two benchmarks, while ID-M$\rightarrow$D7 provides the strongest performance among the evaluated variants, matching or surpassing the original mean at most operating points.

The full IC$\rightarrow$ID-M$\rightarrow$D7 pipeline also maintains strong overall performance, although its results are less consistent at the lowest FARs.

Taken together, these results provide additional evidence that the observed training-efficiency gains extend beyond the validation benchmarks.

\section{Conclusion}


Our work addresses a relatively underexplored aspect of large-scale face-recognition datasets: the presence of distinct training-data defects and their effects on model training. To the best of our knowledge, we provide one of the first controlled analyses comparing how different defect types affect both recognition performance and training efficiency. We also introduce complementary auditing strategies to identify duplicate images, split identities, crop errors, multiple-face ambiguity, and invalid content.

Our results show that these defects remain in MS1MV3 and that their effects on model training differ substantially. Targeted duplicate removal provides the clearest benefit, with the best duplicate-removal setting removing $15.44\%$ of the training images, improving average validation accuracy, and reaching the baseline performance with $23.7\%$ fewer training steps. In contrast, the matched random-removal control does not reproduce these gains, indicating that they do not result from dataset reduction alone. Correcting split identities through label merging and removing invalid content also reduce the number of training steps required to reach baseline performance, while filtering all image-validity defects does not provide the same benefit. Combining split-identity merging with duplicate removal produces the largest reduction in training steps, reaching the baseline with $24.3\%$ fewer steps while maintaining comparable recognition performance.

The strongest variants also remain competitive with the baseline on the held-out IJB-B and IJB-C test benchmarks when evaluated at earlier checkpoints, providing further support for the observed training-efficiency gains. Overall, our findings show that the effect of dataset cleaning is strongly defect-dependent. Removing more images or correcting more defect categories does not necessarily produce a better training set. Instead, identifying and correcting the defects that meaningfully affect training can provide a more effective path toward improving large-scale face-recognition datasets.

{
    \small
    \bibliographystyle{ieeenat_fullname}
    \bibliography{main}

@String(CVPR= {IEEE Conf. Comput. Vis. Pattern Recog.})

@String(CVPR  = {CVPR})

@inproceedings{ms1m,
  title={Ms-celeb-1m: A dataset and benchmark for large-scale face recognition},
  author={Guo, Yandong and Zhang, Lei and Hu, Yuxiao and He, Xiaodong and Gao, Jianfeng},
  booktitle={Proceedings of the European Conference on Computer Vision},
  pages={87--102},
  year={2016},
}

@inproceedings{vggface2,
  title={Vggface2: A dataset for recognising faces across pose and age},
  author={Cao, Qiong and Shen, Li and Xie, Weidi and Parkhi, Omkar M and Zisserman, Andrew},
  booktitle={Proceedings of the IEEE International Conference on Automatic Face \& Gesture Recognition},
  pages={67--74},
  year={2018},
}

@inproceedings{webface260m,
  title={Webface260m: A benchmark unveiling the power of million-scale deep face recognition},
  author={Zhu, Zheng and Huang, Guan and Deng, Jiankang and Ye, Yun and Huang, Junjie and Chen, Xinze and Zhu, Jiagang and Yang, Tian and Lu, Jiwen and Du, Dalong and others},
  booktitle={Proceedings of the IEEE/CVF Conference on Computer Vision and Pattern Recognition},
  pages={10487--10497},
  year={2021},
}

@inproceedings{globallocal,
  title={Global-local gcn: Large-scale label noise cleansing for face recognition},
  author={Zhang, Yaobin and Deng, Weihong and Wang, Mei and Hu, Jiani and Li, Xian and Zhao, Dongyue and Wen, Dongchao},
  booktitle={Proceedings of the IEEE/CVF Conference on Computer Vision and Pattern Recognition},
  pages={7728--7737},
  year={2020},
}

@inproceedings{adaptativenoise,
  title={Adaptive label noise cleaning with meta-supervision for deep face recognition},
  author={Zhang, Yaobin and Deng, Weihong and Zhong, Yaoyao and Hu, Jiani and Li, Xian and Zhao, Dongyue and Wen, Dongchao},
  booktitle={Proceedings of the IEEE/CVF International Conference on Computer Vision},
  pages={15045--15055},
  year={2021},
}

@inproceedings{devilnoise,
  title={The devil of face recognition is in the noise},
  author={Wang, Fei and Chen, Liren and Li, Cheng and Huang, Shiyao and Chen, Yanjie and Qian, Chen and Loy, Chen Change},
  booktitle={Proceedings of the European Conference on Computer Vision},
  pages={780--795},
  year={2018},
}

@inproceedings{serfiq,
  title={SER-FIQ: Unsupervised estimation of face image quality based on stochastic embedding robustness},
  author={Terhorst, Philipp and Kolf, Jan Niklas and Damer, Naser and Kirchbuchner, Florian and Kuijper, Arjan},
  booktitle={Proceedings of the IEEE/CVF Conference on Computer Vision and Pattern Recognition},
  pages={5651--5660},
  year={2020}
}

@inproceedings{magface,
  title={Magface: A universal representation for face recognition and quality assessment},
  author={Meng, Qiang and Zhao, Shichao and Huang, Zhida and Zhou, Feng},
  booktitle={Proceedings of the IEEE/CVF conference on computer vision and pattern recognition},
  pages={14220--14229},
  year={2021},
}

@inproceedings{crfiqa,
  title={{CR-FIQA}: face image quality assessment by learning sample relative classifiability},
  author={Boutros, Fadi and Fang, Meiling and Klemt, Marcel and Fu, Biying and Damer, Naser},
  booktitle={Proceedings of the IEEE/CVF Conference on Computer Vision and Pattern Recognition},
  pages={5836--5845},
  year={2023},
}

@inproceedings{comining,
  title={Co-mining: Deep face recognition with noisy labels},
  author={Wang, Xiaobo and Wang, Shuo and Wang, Jun and Shi, Hailin and Mei, Tao},
  booktitle={Proceedings of the IEEE/CVF International Conference on Computer Vision},
  pages={9358--9367},
  year={2019}
}

@inproceedings{subcenter_arcface,
  title={Sub-center arcface: Boosting face recognition by large-scale noisy web faces},
  author={Deng, Jiankang and Guo, Jia and Liu, Tongliang and Gong, Mingming and Zafeiriou, Stefanos},
  booktitle={Proceedings of the European Conference on Computer Vision},
  pages={741--757},
  year={2020},
}

@inproceedings{rethinking,
  title={Rethinking Robust Representation Learning Under Fine-grained Noisy Faces},
  author={Ma, Bingqi and Song, Guanglu and Liu, Boxiao and Liu, Yu},
  booktitle={Proceedings of the European Conference on Computer Vision},
  pages={614--630},
  year={2022},
}

@inproceedings{doubletrouble,
   title={Double Trouble? Impact and Detection of Duplicates in Face Image Datasets},
   booktitle={Proceedings of the International Conference on Pattern Recognition Applications and Methods},
   author={Schlett, Torsten and Rathgeb, Christian and Tapia, Juan and Busch, Christoph},
   year={2024},
   pages={801–808} }

@misc{facenms,
      title={{Face-NMS}: A Core-set Selection Approach for Efficient Face Recognition}, 
      author={Yunze Chen and Junjie Huang and Jiagang Zhu and Zheng Zhu and Tian Yang and Guan Huang and Dalong Du},
      year={2021},
      eprint={2109.04698},
      archivePrefix={arXiv},
      primaryClass={cs.CV},
}

@misc{diffprob,
      title={{DiffProb}: Data Pruning for Face Recognition}, 
      author={Eduarda Caldeira and Jan Niklas Kolf and Naser Damer and Fadi Boutros},
      year={2025},
      eprint={2505.15272},
      archivePrefix={arXiv},
      primaryClass={cs.CV},
}

@INPROCEEDINGS{datacleaning,
  author={Ng, Hong-Wei and Winkler, Stefan},
  booktitle={Proceedings of the IEEE International Conference on Image Processing}, 
  title={A data-driven approach to cleaning large face datasets}, 
  year={2014},
  volume={},
  number={},
  pages={343-347}}

@INPROCEEDINGS{noisetolerant,
  author={Hu, Wei and Huang, Yangyu and Zhang, Fan and Li, Ruirui},
  booktitle={Proceedings of the IEEE/CVF Conference on Computer Vision and Pattern Recognition (CVPR)}, 
  title={Noise-Tolerant Paradigm for Training Face Recognition {CNNs}}, 
  year={2019},
  volume={},
  number={},
  pages={11879-11888},}

@misc{imagededup,
  title={Imagededup},
  author={Tanuj Jain and Christopher Lennan and Zubin John and Dat Tran},
  year={2019},
  howpublished={\url{https://github.com/idealo/imagededup}},
}

@article{HNSW,
  title={Efficient and robust approximate nearest neighbor search using hierarchical navigable small world graphs},
  author={Malkov, Yu A and Yashunin, Dmitry A},
  journal={IEEE Transactions on Pattern Analysis and Machine Intelligence},
  volume={42},
  number={4},
  pages={824--836},
  year={2018},
}

@inproceedings{BLIP,
  title={Blip: Bootstrapping language-image pre-training for unified vision-language understanding and generation},
  author={Li, Junnan and Li, Dongxu and Xiong, Caiming and Hoi, Steven},
  booktitle={Proceedings of the International Conference on Machine Learning},
  pages={12888--12900},
  year={2022},
}

@inproceedings{arcface,
  title={Arcface: Additive angular margin loss for deep face recognition},
  author={Deng, Jiankang and Guo, Jia and Xue, Niannan and Zafeiriou, Stefanos},
  booktitle={Proceedings of the IEEE/CVF conference on computer vision and pattern recognition},
  pages={4690--4699},
  year={2019}
}

@inproceedings{lfw,
  title={Labeled faces in the wild: A database for studying face recognition in unconstrained environments},
  author={Huang, Gary B and Mattar, Marwan and Berg, Tamara and Learned-Miller, Eric},
  booktitle={Workshop on faces in'Real-Life'Images: detection, alignment, and recognition},
  year={2008}
}

@article{calfw,
  title={Cross-age lfw: A database for studying cross-age face recognition in unconstrained environments},
  author={Zheng, Tianyue and Deng, Weihong and Hu, Jiani},
  journal={arXiv preprint arXiv:1708.08197},
  year={2017}
}

@article{cplfw,
  title={Cross-pose lfw: A database for studying cross-pose face recognition in unconstrained environments},
  author={Zheng, Tianyue and Deng, Weihong},
  journal={Beijing University of Posts and Telecommunications, Tech. Rep},
  volume={5},
  number={7},
  pages={5},
  year={2018}
}

@inproceedings{agedb,
  title={{AgeDB}: the first manually collected, in-the-wild age database},
  author={Moschoglou, Stylianos and Papaioannou, Athanasios and Sagonas, Christos and Deng, Jiankang and Kotsia, Irene and Zafeiriou, Stefanos},
  booktitle={Proceedings of the IEEE Conference on Computer Vision and Pattern Recognition Workshops},
  pages={1997--2005},
  year={2017},
}

@inproceedings{cfp,
  title={Frontal to profile face verification in the wild},
  author={Sengupta, Soumyadip and Chen, Jun-Cheng and Castillo, Carlos and Patel, Vishal M and Chellappa, Rama and Jacobs, David W},
  booktitle= {Proceedings of the IEEE Winter Conference on Applications of Computer Vision},
  pages={1--9},
  year={2016},
}

@inproceedings{IJB,
  title={Iarpa janus benchmark-c: Face dataset and protocol},
  author={Maze, Brianna and Adams, Jocelyn and Duncan, James A and Kalka, Nathan and Miller, Tim and Otto, Charles and Jain, Anil K and Niggel, W Tyler and Anderson, Janet and Cheney, Jordan and others},
  booktitle={Proceedings of the International Conference on Biometrics},
  pages={158--165},
  year={2018},
}

@article{ofiq,
  title={Open source face image quality ({OFIQ})- {Implementation and Evaluation of Algorithms}},
  author={Merkle, J and Rathgeb, C and Herdeanu, B and Tams, B and Lou, D and D{\"o}rsch, A and Schaubert, M and Dehen, J and Chen, L and Yin, X and others},
  journal={Technical report},
  year={2024},
  publisher={Federal Office for Information Security/pn>}
}

@inproceedings{ms1mv3,
  title={Lightweight face recognition challenge},
  author={Deng, Jiankang and Guo, Jia and Zhang, Debing and Deng, Yafeng and Lu, Xiangju and Shi, Song},
  booktitle={Proceedings of the International Conference on Computer Vision Workshop},
  pages={2638--2646},
  year={2019},
}

@misc{insightface,
  author       = {{InsightFace Project}},
  title        = {InsightFace: 2D and 3D Face Analysis Project},
  howpublished = {\url{https://github.com/deepinsight/insightface}},
}

@inproceedings{faceqnet,
  title={{FaceQnet}: Quality assessment for face recognition based on deep learning},
  author={Hernandez-Ortega, Javier and Galbally, Javier and Fierrez, Julian and Haraksim, Rudolf and Beslay, Laurent},
  booktitle={Proceedings of the International Conference on Biometrics},
  pages={1--8},
  year={2019},
}

@inproceedings{adaface,
  title={Adaface: Quality adaptive margin for face recognition},
  author={Kim, Minchul and Jain, Anil K and Liu, Xiaoming},
  booktitle={Proceedings of the IEEE/CVF Conferencee on Computer Vision and Pattern Recognition},
  pages={18729--18738},
  year={2022},
}
}

\end{document}